\documentclass[epsfig]{bmcart}

\usepackage{epsfig}

\usepackage{url}
\usepackage{cite}
\usepackage{subfigure}
\usepackage{subfig}
\usepackage{graphics}
\usepackage{float}
\usepackage{amsthm,amsmath}
\usepackage[utf8]{inputenc} 

\usepackage[aboveskip=1pt,labelfont=bf,labelsep=period,justification=raggedright,singlelinecheck=off]{caption}
\usepackage{epstopdf}
\usepackage{lastpage,fancyhdr,graphicx}

\startlocaldefs

\endlocaldefs

\begin{document}

\begin{frontmatter}

\begin{fmbox}
\dochead{Research}


\title{AMFPMC - An improved method of detecting multiple types of drug-drug interactions using only known drug-drug interactions}


\author[
  addressref={aff1},                   
  corref={aff1},                       
  email={veredba@post.bgu.ac.il}   
]{\inits{B.V.}\fnm{Bar} \snm{Vered}}
\author[
  addressref={aff1,aff2},
]{\inits{L.R}\fnm{Guy} \snm{Shtar}}
\author[
  addressref={aff1,aff2},
]{\inits{L.R}\fnm{Lior} \snm{Rokach}}
\author[
  addressref={aff1,aff2},
]{\inits{G.S}\fnm{Bracha} \snm{Shapira}}


\address[id=aff1]{
  \orgdiv{Department of Software and Information Systems Engineering},             
  \orgname{Ben-Gurion University of the Negev},          
  \city{Beer-Sheva},                              
  \cny{Israel}                                    
}



\end{fmbox}


\begin{abstractbox}

\begin{abstract} 
\parttitle{Background} 
Adverse drug interactions are largely preventable causes of medical accidents, which frequently result in physician and emergency room encounters. The detection of drug interactions in a lab, prior to a drug's use in medical practice, is essential, however it is costly and time-consuming. Machine learning techniques can provide an efficient and accurate means of predicting possible drug-drug interactions and combat the growing problem of adverse drug interactions. Most existing models for predicting interactions rely on the chemical properties of drugs. While such models can be accurate, the required properties are not always available.

\parttitle{Results} 
In this article we address the drug-drug interaction issue as a link prediction problem and extend a method proposed by Shtar et al~\cite{binaryDDI}, which uses artificial neural networks and propagation over graph nodes in order to consider specific interactions when detecting drug-drug interactions. After extracting and analyzing the possible interactions, a table which presents the interactions as a one-hot vector between each pair of drugs is created. Then, a deep neural network (DNN) is used as a predictor.  in the training stage, receiving two vectors in which the corresponding indexes are two drugs with a specific interaction, and then, predicting an interaction for two drugs with an unknown interaction. We perform holdout and retrospective analyses using DrugBank data. Our results show that the proposed graph similarity derived method, which is a graph similarity derived method, outperforms models that use chemical and biologic properties and other state-of-the-art models and is very efficient, as it uses less data to generate better predictions in less time.
\parttitle{Conclusion} 
In this research, we have extended our DDI prediction algorithm while using the same graph similarities and artificial intelligence techniques. This extension enabled the model to detect a specific interaction and not only predicting whether there is or there isn’t an interaction between a pair of drugs. The drug-drug interaction prediction problem should be solved using different and diverse datasets, in AMFPMC, these diverse and different datasets are important since our algorithm relays on interactions network, and interactions network are not helpful when it comes to new drugs with unknown interactions. The main difference between many studies and our research is that usually these studies relay on a multiple domain dataset, which means they use several chemical properties of drugs while our algorithm solely relays on basic mathematical properties. This usually means that our algorithm and DNN’s input used are quite simpler and enabling future extensions. Our evaluation and results demonstrates the proposed method’s superiority, which can be seen in the metrics calculated comparison. Moreover, we demonstrate that the embedding of the input to the DNN can be used and be helpful in other various problems. \par

\end{abstract}




\end{abstractbox}
%

\end{frontmatter}
\clearpage




\section*{Background}
Adverse drug events are considered medical injuries and are thought to be one of the top causes of death in the U.S., ahead of many chronic diseases such as diabetes, AIDS, and heart disease~\cite{FDA_drug_interactions}. The cost attributed to these reactions is high and is estimated at thousands of dollars per patient annually in the U.S.~\cite{doi:10.1001/jama.1997.03540280045032}. The range of patients that are harmed by drug-drug interactions range 3-5\% of all medications errors within hospitals. Furthermore, ADRs (adverse drug reactions) represent a substantial proportion of the number of patients seeking medical care in general and the number of patients rushed to intensive care units~\cite{pmid10192758, pmid17047216} in particular.
\par

Many 
studies 
using computational techniques for the detection of DDIs have been published in the last decade. In most cases, initial efforts focused on detecting interactions in a binary manner, using an algorithm to detect whether an interaction between two drugs exists or not. However, recent studies have proposed models capable of detecting a specific set of interactions. Various similarity-based models have been proposed, including models based on DDI similarities~\cite{10.1371/journal.pone.0058321}, side effect similarities~\cite{Zhang2015}, structural similarities~\cite{RyuE4304}, and a combination of similarities~\cite{Zhang2017, 10.1371/journal.pone.0140816, Gottlieb592}. In one study natural language processing techniques were used to embed the interactions and then the embeddings were used to predict DDIs~\cite{10.1371/journal.pone.0190926}, while another used integrated similarity and then applied another neural network to classify the interaction~\cite{narjesNature}; In the latter, the problem is firstly considered a binary problem in which the model detects whether there is an interaction. After this stage the output generated by the model served as an input to an additional model which predicted a specific interaction. This way the model's complexity is divided between two neural networks. There are models which use the interactions of other components to predict DDIs; for example, in one proposed method substructure-substructure interactions were used~\cite{10.1093/bib/bbab133}; that method is similar to the one proposed in this study, as it utilizes a graph and graph similarities techniques, but the two models differ in that the authors of that study used a graph and graph similarities techniques on substructures while we use it on matrices. In a more complex method, other properties in addition to chemical substructures were used (e.g., biologic properties), enabling the prediction of interactions between biologic drugs and other types of drugs~\cite{10.1186}.
\par

\begin{figure}[h!]
\includegraphics[width=\textwidth]{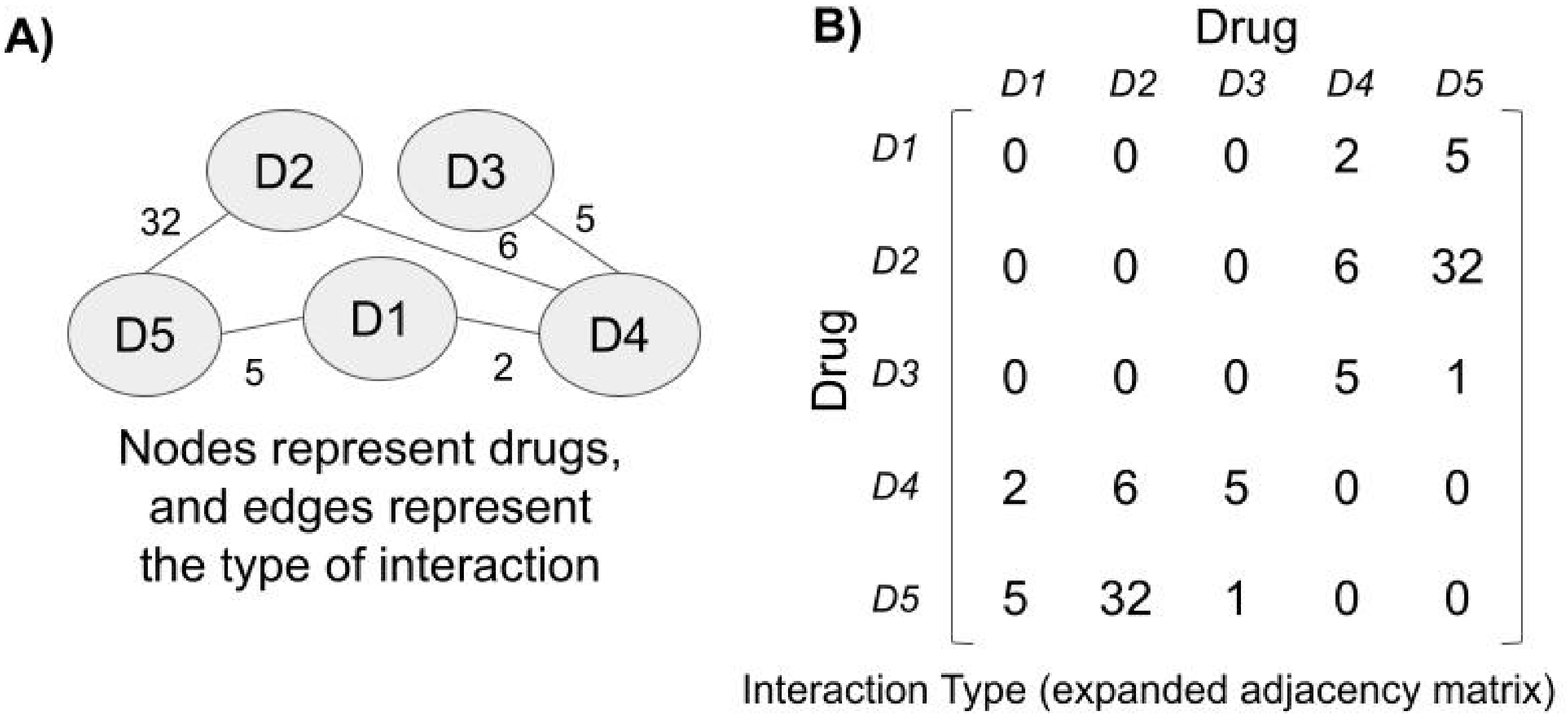}
\caption{{Link prediction problem reduction of the DDI problem.} \textbf{A)} A DDI graph \textbf{B)} The DDI graph is represented by an adjacency matrix: the rows and columns represent drugs, and the different values stand for different interactions. For example, the cell in the first row and the right-most column represents the interaction between D1 and D5. In a link prediction problem, a score is calculated for every non-existent interaction.}
\label{fig0}
\end{figure}

DDI detection using an interaction network can be reduced to a link prediction problem in a graph, this reduction is used in this study. In a link prediction problem, the aim is to accurately detect and predict the edges (interactions) between the nodes (drugs) that will be inserted into the graph. The most intuitive approach for this is to add edges to nodes for which there is a large overlap; this means that if nodes A and B have many similar neighbors, and they are not currently connected by an edge, they might need to be connected because of the large overlap between their neighbors; In the case of DDIs, edges are interactions, and nodes are drugs, and when two drugs have many similar neighbors, there may be  an interaction between them. The DDI prediction problem involves a different characterization of the link prediction problem in which edges indicate whether an interaction occurs between two drugs (two nodes); here, we want to be more specific by indicating a type of edge that corresponds to a specific interaction. This problem can be approached as a link prediction in weighted multiplex systems~\cite{Shikhar} problem. Another approach, which is less intuitive but more powerful, is the matrix factorization method in which a matrix is factorized into a product of matrices. This technique is widely used in dimensionality reduction area, and recently, many studies have successfully factorized a matrix using deep neural networks~\cite{He:2017:NCF:3038912.3052569, WU201846, FAN201834}. Figure~\ref{fig0} present a Link prediction reduction example of the adjacency matrix representing the DDI graph. Although this technique has obtained meaningful results in the DDI field, it has some drawbacks, such as asymmetric decomposition when the matrix's transpose should be equal to the original matrix; this occurs because the row vectors and column vectors are identical. Another drawback is the need to limit the score to the range [0,1].
\par

For an accurate DDI prediction, a neural network that encompasses the linear structure of the interaction graph is required. In this study, AMFPMC receives a one-hot encoding representation of two nodes as input. The model's output is a single class representing the interaction between the two drugs. In AMFPMC, the numbers greater than zero indicate specific types of interactions, the value of zero has different meanings according to the evaluation used (retrospective or holdout). Only the drug interaction graph is needed to utilize this model, and no other domain-specific information (such as chemical properties and substructures) is required.

\begin{figure}[h!]
\includegraphics[width=\textwidth]{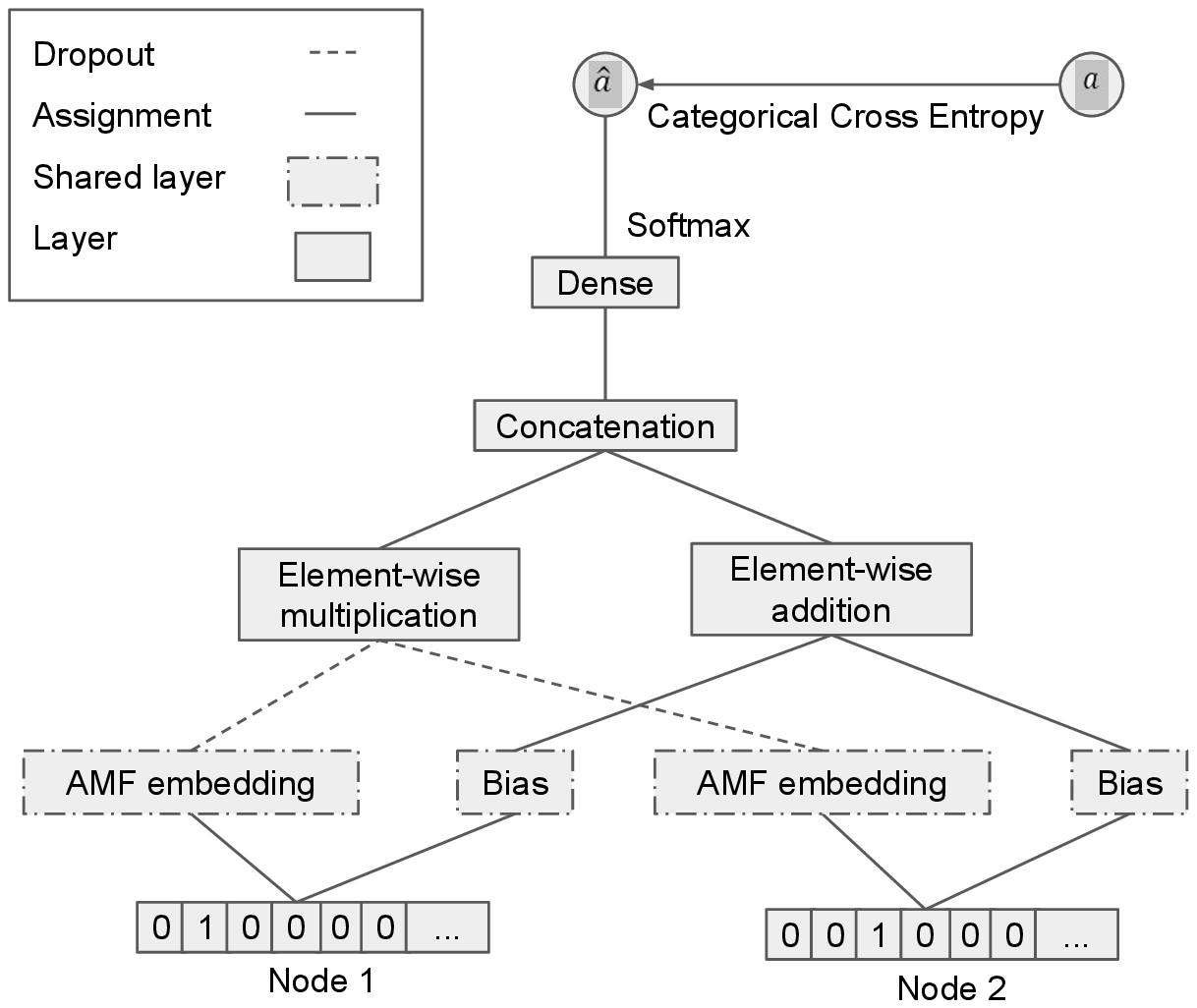}
\caption{{\bf Overview of AMFPMC's architecture.}
Drugs are represented as nodes; embedding layers (which act as latent factors) and biases are shared between input nodes. Dropout is used as a regularization mechanism to prevent overfitting.}
\label{fig1}
\end{figure}

\clearpage

\section*{Evaluation}
In this section, we discuss the two experiments performed to properly evaluate AMFPMC (AMFPMC): retrospective and holdout evaluations using drugs from various versions of the DrugBank database~\cite{pmid29126136}. We have also evaluated AMFPMC on specifically biologic drugs and checked its performance, this is due to prior studies often neglecting biologic drugs and supporting just small molecules. In addition, we have evaluated the performance of the embedding created by the model further, to assess its ability to address other problems (besides drug interaction prediction) and determine whether it contains useful information about drugs. \par
\begin{figure}[h]

\includegraphics[width=\textwidth]{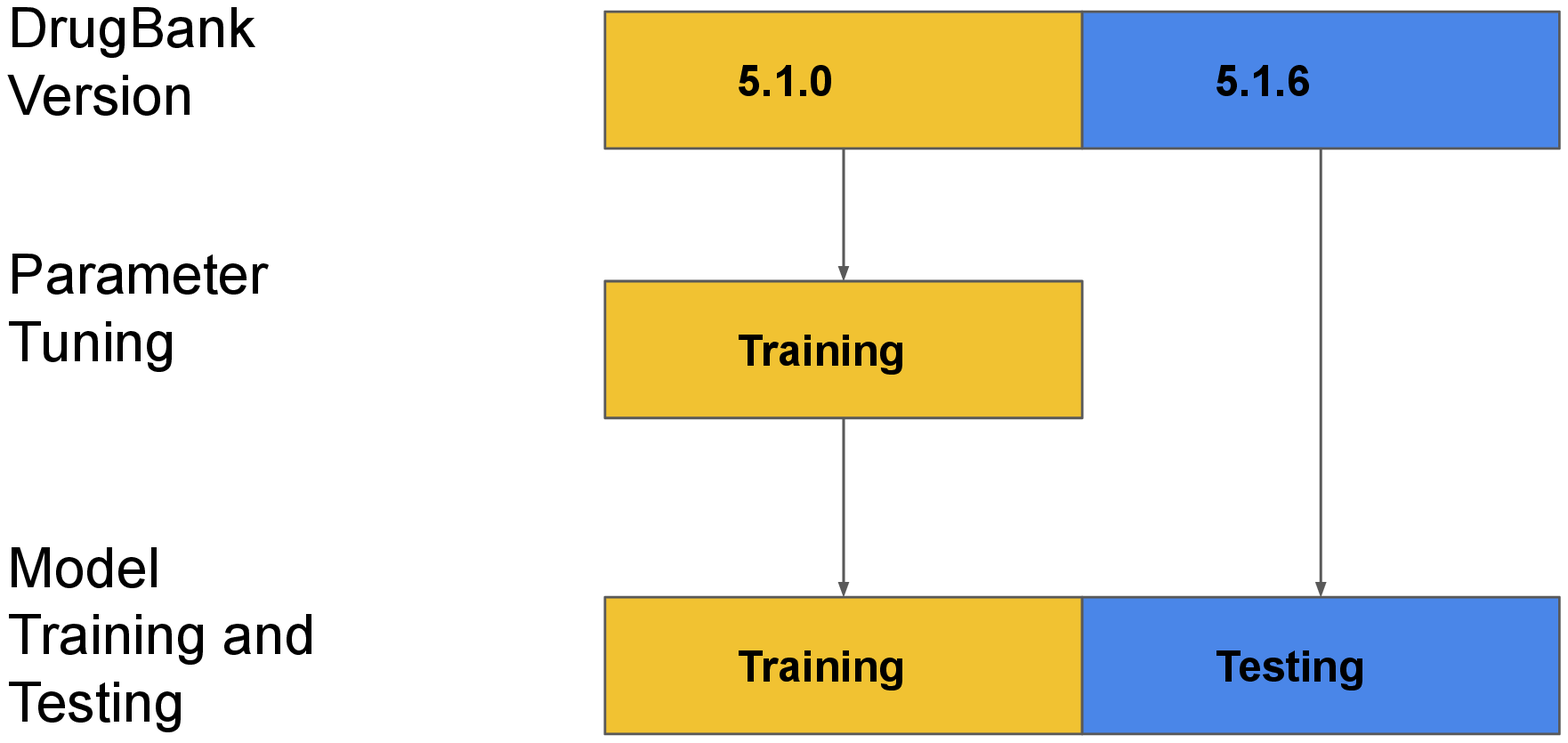}
\caption{{\bf Training and testing evaluation for in retrospective evaluation.} In the training stage, parameter tuning is performed using DrugBank version $5.1.0$. The final model is trained using these parameters with the data from DrugBank version $5.1.0$ and tested using DrugBank version $5.1.6$.}
\label{fig2}
\end{figure}

Figure~\ref{fig2} presents the training and testing evaluation for the retrospective evaluation in which two version of the DrugBank database were used. There are many differences between the two versions - many new interactions were added to the more recent version, along with some modifications to interactions included in the earlier version (which can interfere the learning process). DrugBank's version $5.1.0$ (published in April $2018$) served as the dataset for training, and DrugBank's version $5.1.6$ (published in April $2020$) was used for testing. Version $5.1.0$ of the database contains $321,738$ interactions and $2,904$ drugs, while version $5.1.6$ contains $1,334,875$ interactions and $4,264$ drugs. To ensure dataset compatibility for the model introduced in this paper. Only the drugs that appear in both versions were extracted during the preprocessing stage. For the retrospective evaluation, a total of $37$ interactions were extracted in the preprocessing stage; this includes interaction numbered $0$ which is no interaction, interactions numbered $1-35$ which are interactions that are considered common (have plenty of pairs of drugs with these interactions), and interactions numbered $36$ which are "other" - a set of uncommon interactions which were grouped together because of their rarity. For the holdout evaluation a total of $60$ interactions were extracted, furthermore, the holdout evaluation uses a modified DrugBank's data and these changes are elaborated below. Since the adjacency matrix is symmetric, we used the original wording in the interactions and extracted the keyword phrases; then we removed words that are redundant, for example, in the case of the following sentence from DrugBank: "$The \ metabolism \ of \ Drug \ b \ can \ be \ decreased \ when \ combined \ with \ Drug \ a$" we extract "$metabolism \  decreased$," and an index was assigned to represent the interaction (as a number) in the adjacency matrix. In this example, the number $2$ was assigned, and the cell representing the interaction between $Drug \ a$ and $Drug \ b$ holds the value $2$. Since all of the interactions are symmetric, the $Drug \ a$ and $Drug \ b$ wording in the interaction, which is comprised of the keyword phrase, is not needed. In the holdout evaluation, the interactions were randomly selected, although in reality some interactions are more likely to be found due to the drugs prevalence in drug combinations, the interactions prevalence in drug combinations, drugs chemical compounds, and more.

\clearpage

In the holdout evaluation, a tuned version of DrugBank $5.1.3$ which was used. This version, which only includes molecule drugs, was compiled by Deng and his team~\cite{deng2020multimodal} and is required for the comparison. This version includes $572$ drugs and $37,264$ interactions among them. In this version, interactions numbered $0-64$ are valid interactions, and there is no representation for non-interactions like there is in the retrospective evaluation. In the holdout evaluation, we have also compared AMFPMC's performance to that of the following state-of-the-art models: random forest (RF)~\cite{RandomForest}, k-nearest neighbors (KNN), and logistic regression (LR).

\subsection*{\textbf{Metrics}}
The following metrics are used to evaluate the performance of the model in a holdout evaluation: the accuracy, area under the receiver operating characteristic curve (AUROC), F1-score, recall, and precision metrics. We also used the area under the precision-recall curve (AUPR) metric, which could be relevant for use in other link prediction problems~\cite{Yang2015}. For the retrospective evaluation, the AUROC served as our main metric, since this evaluation is mainly aimed at demonstrating AMFPMC's ability to predict biologic drugs.
\subsection*{\textbf{Baselines}}
We compared our AMFPMC method to the following models:
\begin{itemize}
 \item
  DDIMDL which was used by Deng et al.~\cite{deng2020multimodal}.
  This method relies on the chemical properties of drugs, such as drug's smiles, enzymes, and targets, it also uses Jaccard similarity, since each vector contains properties about the drugs themselves, rather than just the index as done in AMFPMC. DDIMDL uses a single database which means it is not suitable for retrospective evaluation. To compare DDIMDL and AMFPMC, as a part of the holdout evaluation we performed $5$-fold cross-validation.
\item 
    State-of-the-art models: RF, LR, and KNN. These models perform well in the scenarios of link prediction. As in the case of the model proposed by Deng et al, these models receive the following features as input: drug's smiles, enzymes, and targets.
\end{itemize}

\par
We implemented AMFPMC using Keras~\cite{chollet2015keras} as was done in DDIMDL. However, their model was configured to be evaluated solely using holdout evaluation, while ours was configured to be evaluated using retrospective evaluation. This is due to their usage of a single database. To enable comparison between the models, we refactored our code so it could run on the same data they used.

\subsection*{\textbf{Parameter tuning}}
In order to determine the optimal parameters for each evaluation, we used grid search to identify the maximum accuracy value on the validation set. We did so by extracting $20$\% of the test data and used this as a validation set. We then trained on the original training set and validated on the validation set. Then, the parameters values which achieved the maximum accuracy values in the grid search were used in AMFPMC to achieve the best results. This was done twice, once for the holdout evaluation and once for the retrospective evaluation. In both cases, we examined the following batch sizes \{$128$, $256$, $512$, $102$4\} and learning rates \{$0.1$, $0.01$, $0.001$, $0.0001$\}, and the following ranges for the  dropout levels [$0$-$0.9$], number of epochs [$1$-$50$], and propagation factors [$0$-$1$] (in intervals of $0.1$). The optimal parameters, which were used in our experiments, are presented in Table ~\ref{table1}.
\begin{table}[h!]
\caption{Optimal parameters for each evaluation: retrospective and holdout}
\begin{tabular}{|| ccccccc ||}
        \hline 
        Evaluation type & Embedding Size & Dropout & Epochs & Batch Size & Learning Rate & Propagation Factor\\ \hline
        Retrospective & 512 & 0.3 & 5 & 1024 & 0.01 & 0.8\\ 
        Holdout &  512 & 0.3 & 15 & 256 & 0.01 & 0.6-1.0 \\\hline
\end{tabular}
\label{table1}
\end{table}

\section*{Results}
In this section, we present AMFPMC's results in the holdout and retrospective evaluations (we also present the results of embedding evaluation). The retrospective evaluation includes the comparison between biologic and molecule drugs. As seen below, the results demonstrate AMFPMC's superiority and ability to outperform the other models on most metrics, particularly the AUROC and AUPR. It can also be seen that AMFPMC performs equally well on biologic drugs due to the fact that it uses general properties, such as interaction neighboring, rather than relying solely on chemical substructures and properties. Our results are mainly based on the holdout evaluation - because we compare AMFPMC to DDIMDL and due to the fact that Deng et al~\cite{deng2020multimodal} provided a dataset that is relevant only to a single version (DrugBank $5.1.3$), this prevents us from performing a retrospective evaluation on their data. Other state-of-the-art models are also compared to in this evaluation.. Therefore the findings presented are based mainly on the holdout evaluation.
\subsection*{\textbf{Holdout evaluation}}
Holdout evaluation was performed by using the tuned version DrugBank's database provided by Deng et al ~\cite{deng2020multimodal}, which is a subset of DrugBank $5.1.3$ that only includes molecule drugs. In AMFPMC original setup, zero indicates no interaction, to enable a holdout evaluation, we modified the DNN so that AMFPMC could be trained, tested, and make predictions. In our comparison, $5$-fold cross-validation was used.

\begin{table}[htp!]
\caption{\bf Performance comparison (micro-comparison and accuracy)} 
    \begin{tabular}{|| ccccccc ||}
    \hline
    Model & ACC & AUPR & AUC & F1-SCORE & PRECISION & RECALL \\
    \hline
     AMFPMC & \textbf{0.8964} & \textbf{0.9565} & \textbf{0.9987} & \textbf{0.8945} & \textbf{0.8944} & \textbf{0.8943} \\
    DDIMDL & 0.8818 & 0.9334 & 0.9977 & 0.8817 & 0.8816 & 0.8815 \\
    KNN & 0.7147 & 0.7793 & 0.9806 & 0.7147 & 0.7146 & 0.7145 \\
    LR & 0.7211 & 0.7845 & 0.9933 & 0.7211 & 0.7210 & 0.7209 \\
    RF & 0.7729 & 0.8457 & 0.9953 & 0.7729 & 0.7728 & 0.7727 \\
    \hline
    \end{tabular}
    \label{table2}
\end{table}

\begin{table}[htp!]
\caption{\bf Performance comparison (macro-comparison)}
    \begin{tabular}{|| cccccc ||}
    \hline
    Model & AUPR & AUC & F1-SCORE & PRECISION & RECALL \\
    \hline
    AMFPMC & \textbf{0.8821} & \textbf{0.9915} & 0.7534 & \textbf{0.8878} & 0.7001 \\
    DDIMDL~\cite{deng2020multimodal} & 0.8343 & 0.9873 & \textbf{0.7706} & 0.8723 & \textbf{0.7271} \\
    KNN & 0.6266 & 0.9299 & 0.4917 & 0.7491 & 0.4125\\
    LR & 0.6115 & 0.9796 & 0.3062 & 0.5036 & 0.2543\\
    RF & 0.6545 & 0.9752 & 0.4871 & 0.7156 & 0.4134 \\
    \hline
    \end{tabular}
    \label{table3}
\end{table}

\begin{table}[htp!]
\caption{\bf Top-$20$ types of interactions keyword phrases along with the model's AUROC in order based on the \# of samples of each interaction, predicted as holdout on DrugBank $5.1.3$.}

  \begin{tabular}{|| cccc ||}
    \hline
    Interaction & AMFPMC & DDIMDL  & \# samples\\ \hline
    the metabolism decreases & \textbf{0.9344}

  &	0.9343

 & 9810\\
    the risk or severity of adverse effects increases & \textbf{0.9576}
 &	0.9513

 & 9496\\ 
    the serum concentration increases & \textbf{0.9289}
 &	0.9028

 & 5646\\ 
    the serum concentration decreases & \textbf{0.9440}
	 & 0.9169

 & 2386\\ 
    the therapeutic efficacy decreases & 0.9084
	& \textbf{0.9188}

 & 1312\\
    the central nervous system depressant ( CNS depressant ) activities increases & \textbf{0.9873}
 & 	0.9645

 & 1132\\
    the QTc - prolonging activities increases & \textbf{0.9309}
 &	0.9140

& 1102\\
    the hypotensive activities increases & \textbf{0.9620}
 &0.9485

 & 1086\\
    the metabolism increases & 0.8464
	& \textbf{0.8790}

 & 695\\ 
    the antihypertensive activities decreases & 0.9336
 &	\textbf{0.9588}

 & 551\\ \
    the hypoglycemic activities increases & \textbf{0.9611}
 &	0.9430

 & 362\\ 
    the anticoagulant activities increases & \textbf{0.9444}
 &	0.9348

 & 318\\ 
    the serum concentration of the active metabolites increases & \textbf{0.8911}
 &	0.7750

 & 245\\ 
    the bradycardic activities increases & \textbf{0.9426} &

0.9201
 
 & 245\\ 
   the serotonergic activities increases & 0.8083
& \textbf{0.8162}

 & 188\\ 
    the therapeutic efficacy increases & 0.8636
	& \textbf{0.8967}

 & 165\\ 
    the hypokalemic activities increases & \textbf{0.9938}
 &	0.9785

& 163\\ 
   the orthostatic hypotensive activities increases & \textbf{0.9244}
 &	0.8738

 & 159\\ 
    the cardiotoxic activities decreases & 0.9523
	& \textbf{0.9648}

 & 158\\ 
    the excretion rate decreases, which could result in a higher serum level & \textbf{0.9643}
& 0.9642

 & 154\\
 \hline
  \end{tabular}
  \label{table4}
\end{table}https://www.overleaf.com/project/6289e22895afb9cfbfd6ea77

\clearpage 
\begin{figure}[h]
\includegraphics[width=\textwidth]{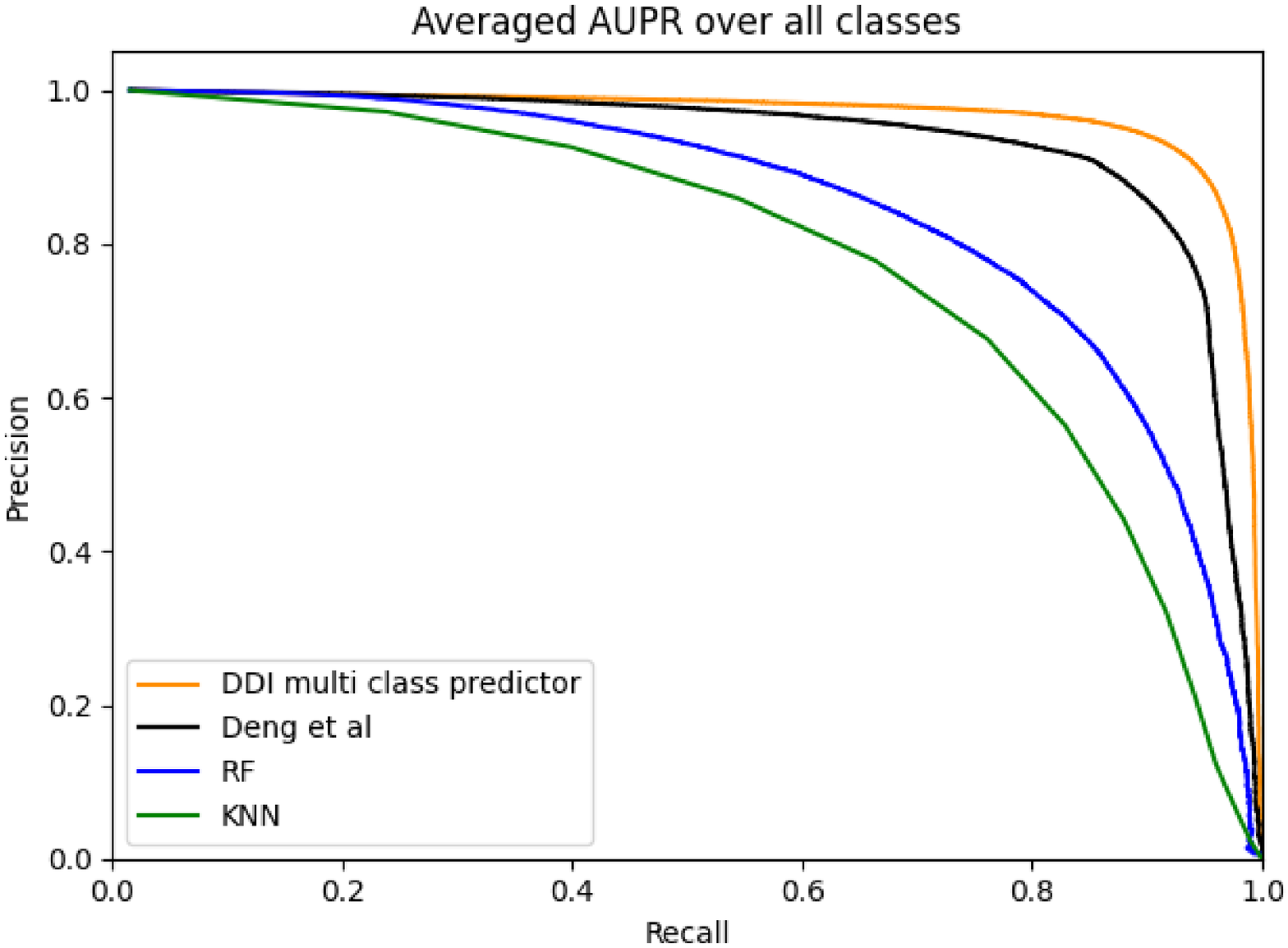}
\caption{{\bf AUPR curves for top-$4$ algorithms in the evaluation.} LR had results that were close to $0.5$ so it is not shown in this graph.}
\label{fig6}
\end{figure}

Table~\ref{table2} presents a comparison of the examined models performance on all of the metrics, with values for each model calculated using micro-calculations and the accuracy scores, while Table~\ref{table3} presents the macro-calculations, with the best scores appearing in bold. As can be seen, AMFPMC outperforms the other models on almost all of the metrics. Figure~\ref{fig6} presents the AUPR values obtained for all of the models (except for the LR model which scored poorly) averaged over all of the interactions classes. A graph of the AUC scores is less useful (and therefore was not included), since the AUC tends to be high in holdout evaluations and the results were quite close to each other. As shown in the figure, AMFPMC is superior to the other models in terms of the AUPR. Table ~\ref{table4} provides a brief summary of the top-$20$ interactions in terms of the number of samples and their AUROC scores for AMFPMC and DDIMDL.

\clearpage

\subsection*{\textbf{Retrospective evaluation}}
In this evaluation, AMFPMC was trained on DrugBank's version $5.1.0$ and tested on DrugBank version $5.1.6$. In this case, the pairs of drugs that the model is trained on are pairs that have some interaction between them, and the pairs of drugs that are tested on are the pairs that have no interaction between them; the final prediction is then verified using DrugBank's version $5.1.6$. This evaluation of AMFPMC is to predict on solely biologic drugs. The point of this evaluation is to evaluate AMFPMC's ability to predict on biologic drugs.

In table 5 the results obtained when AMFPMC is only used to predict interactions between biologic drugs are shown.

\begin{table}[htp!]
\caption{\bf AMFPMC's performance when predicting interactions between biologic drugs }
    \begin{tabular}{|| ccccc ||}
    \hline
    & \# of Drugs & \# of DDIs & AUC & AUPR \\
    \hline
    Biologic Drugs & 452 & 162,949 & 0.820 & 0.98\\
    \hline
    \end{tabular}
\end{table}

\subsection*{\textbf{Embedding evaluation}}
As can be seen in Figure~\ref{fig1}, AMFPMC creates an embedding for each drug by performing various functions on the input data. This embedding contains useful data about the drug which can be used in drug interaction prediction as well as other areas. To evaluate the embedding's capability for this, we used two datasets and added a column containing an embedding for each drug in each dataset; then we used state-of-the-art models to examine whether the data in this column was could predict the result column. We used datasets from a paper on drug safety in pregnancy~\cite{GuyPreg} and an unpublished paper on the prediction of drugs' anticancer activity. In those papers, the model was used to predict respectively whether a drug is safe to use within being pregnant and or has anticancer activities. Tables 6 and 7 compare the performance of AMFPMC and AMFP (a binary model) in terms of the AUC. As can be seen, despite the fact that AMFPMC and its embeddings are quite complex, AMFPMC's performance still often surpasses the performance of AMFP on each dataset; these results demonstrate that AMFPMC's embeddings have widespread relevance and can be used for problems beyond drug interaction prediction.

\begin{table}[htp!]
\caption{\bf Performance (AUC) of AMFPMC and AMFP embeddings in the pregnancy drug safety task}
    \begin{tabular}{|| ccc ||}
    \hline
    & AMFPMC  & AMFP  \\
    \hline
    XGBoost & 0.676 & 0.586\\
    Random Forest & 0.702 & 0.689\\
    LightGBM & 0.7 & 0.620\\
    \hline
    \end{tabular}
\end{table}

\begin{table}[htp!]
\caption{\bf Model's embeddings result in anticancer activities drugs}
    \begin{tabular}{|| ccc ||}
    \hline
    & AMFPMC AUC & AMFP AUC \\
        \hline

    XGBoost & 0.915 & 0.9133\\
    Random Forest & 0.913 & 0.909\\
    LightGBM & 0.901 & 0.908\\
    \hline
    \end{tabular}
\end{table}

\section*{Case studies}
In this section, we analyze the performance of AMFPMC by taking the predictions of AMFPMC on unobserved interactions of pairs of drugs that were extracted from DrugBank's version $5.1.9$ (the most recent version at the time of this study. Then, we check on various validation sources to see if those sources also mention or hint at the drugs interaction. More specifically, for interactions for which AMFPMC has a high confidence in its prediction, we try to obtain some kind of confirmation or hint from another source that corroborates our finding and helps us better understand AMFPMC's decision. Table 8 includes the top-10 interactions predicted by AMFPMC ranked by the model confidence. This section is highly important as it emphasizes the strength of DDI prediction models.
To perform this evaluation, the $40$ interactions for which our model had the greatest confidence in its prediction were extracted. Of these $40$, we identified eight interactions which did not appear in DrugBank's version $5.1.9$ which were mentioned in various other sources (e.g., the literature, drug manuals). We note that DrugBank is updated on a daily basis, and these interactions will likely be added to the next version released.
Table~\ref{table8} provides additional details pertaining to our evaluation of these eight interactions.

\begin{table}[htp!]
\caption{\bf $8$ interactions that were not observed in DrugBank $5.1.9$ (the most recent version at the time of this study) but were predicted by AMFPMC and also either confirmed or hinted at by various other sources.}
    \begin{tabular}{|| p{0.3cm} p{1.6cm} p{1.8cm} p{1.5cm} p{4cm} p{4cm}||}
    \hline
    \# & Drug 1 & Drug 2 & Signaling Source & Interaction Predicted  & Explanation\\
    \hline
    1 & Lixisenatide & Insulin Glargine & Wikipedia & Lixisenatide may increase the hypoglycemic activities of Insulin Glargine. & According to Wikipedia~\cite{wikiinsulin}, this drug combination can cause hypoglycemia. \\
    \hline
    2 & Oxitriptan & Melatonin & DrugBank & The risk or severity of adverse effects can be increased when Oxitriptan is combined with Melatonin. & According to DrugBank~\cite{5-HTP-page}, Oxitriptan can cause increased Serotonin production; since Serotonin is a precursor for Melatonin, one can consider that taking the two drugs together could result in an overdose. \\
    \hline
    3 & Docusate & Phenolphtalein & NCBI & 	The risk or severity of adverse effects can be increased when Docusate is combined with Phenolphtalein. & Some studies~\cite{docu-phenol} indicate adverse effects when the drugs are taken together. \\ \hline
    4 & Chromic Chloride & Cyanocobalamin & NCBI & Chromic Chloride may decrease the excretion rate of Cyanocobalamin which could result in a higher serum level. & According to an NCBI article~\cite{excretionCyano}, this combination of drugs could result in decreased excretion and increased efficiency of Cyancobalamin. \\
    \hline
    5 & Bromazepam & Propantheline  & ndrugs & Bromazepam may increase the CNS depressant activities of Propantheline. & According to a manual~\cite{broma-propa-manual} which relies on references from European health department, this drug combination is not recommended for patients with pre-existing CNS depression. \\
    \hline
    6 & Tocopherol & Cyanocobalamin & NCBI & Tcopherol may decrease the excretion rate of Cyanocobalamin which could result in a higher serum level.  & ccording to an NCBI article~\cite{toco-cyano} hinted about a probable interaction between the two related to the excretion rate. \\
    \hline
    7 & Isopropamide & Trifluoperazine & TabletWise & Isopropamide may increase the CNS depressant activities of Trifluoperazine. & According to a manual~\cite{iso-triflu} an individual should not take this drug combination if he/she has pre-existing CNS depression. \\
    \hline
    8 & Clidinium & Chlordiazepoxide & NIH DailyMed & Clidinium may increase the CNS depressant activities of Chlordiazepoxide. & In the manual~\cite{dailymed} there is an entire paragraph  discussing adverse CNS reactions seen when taking this drug combination. \\ 
    \hline

    \end{tabular}
    \label{table8}
\end{table}

\clearpage

\section*{Discussion}
Drug interactions are the cause of many emergency room visits. Estimates of the percentage of patients harmed by drug interactions range from $3$-$5$\% of all medication errors within hospitals~\cite{pmid10192758, pmid17047216}. In recent years, the adoption of drug-drug interaction detection models and prediction models has increased, particularly during the COVID-19 pandemic~\cite{10.1001/jamanetworkopen.2022.7970}. Drug-drug interactions are usually not identified until late in the research and clinical stages; therefore, drug-drug interaction models are still considered the most practical way of identifying potentially harmful interactions since they can be performed earlier in the drug experimentation process. In our latest article we introduced AMF and AMFP, two models for DDI detection; we also used DrugBank's powerful databases to compare our proposed models to other models, a comparison which demonstrated the new models superiority on the AUROC, AUPR, accuracy, and F1-score, precision, and recall metrics. That study focused on predicting whether an unobserved interaction exists or not, while the current research empowers these models with the ability to predict the type of interaction present rather than just whether an interaction exists. Our proposed model can handle multiple types of DDIs, and our findings confirm that an interaction network of DDIs is the most useful data source for discovering previously unobserved DDIs. The models flexibility is due to the fact that AMFPMC does not rely on knowledge of drugs chemical structures or properties and instead relies on graph similarities. Moreover, AMFPMC also has the ability to predict interactions between biologic drugs. In addition, since it relies on a very small set of properties (just the interaction network), it provides accurate results quickly and effciently. The use of weight balancing was also shown to be beneficial to AMFPMC, especially on small datasets or datasets which have interactions that have a very small number of samples (i.e., less than 10), and this finding demonstrates AMFPMC's effectiveness on datasets of various sizes.
\par
We performed an evaluation of the AUROC score as a function of the propagation factor, both in the holdout and retrospective evaluations. For the holdout evaluation, the optimal values of the propagation factor were [$0.6$ - $1$], and for the retrospective evaluation, the optimal values were $0.6$ and $0.7$. The retrospective evaluation is of course preferable, because it is more realistic and indicates whether a model truly generalizes or not. 
The holdout evaluation is helpful when comparing the results of multiple models, as it can show which model produces better results (on average), while for case studies and predicting unobserved interactions, a retrospective evaluation is preferable. As can be seen, the fluctuations in the holdout evaluation metrics comparisons are small, and in the retrospective evaluation, the propagation factor has more impact.
\par
\begin{figure}[h!]
\includegraphics[width=0.8\textwidth]{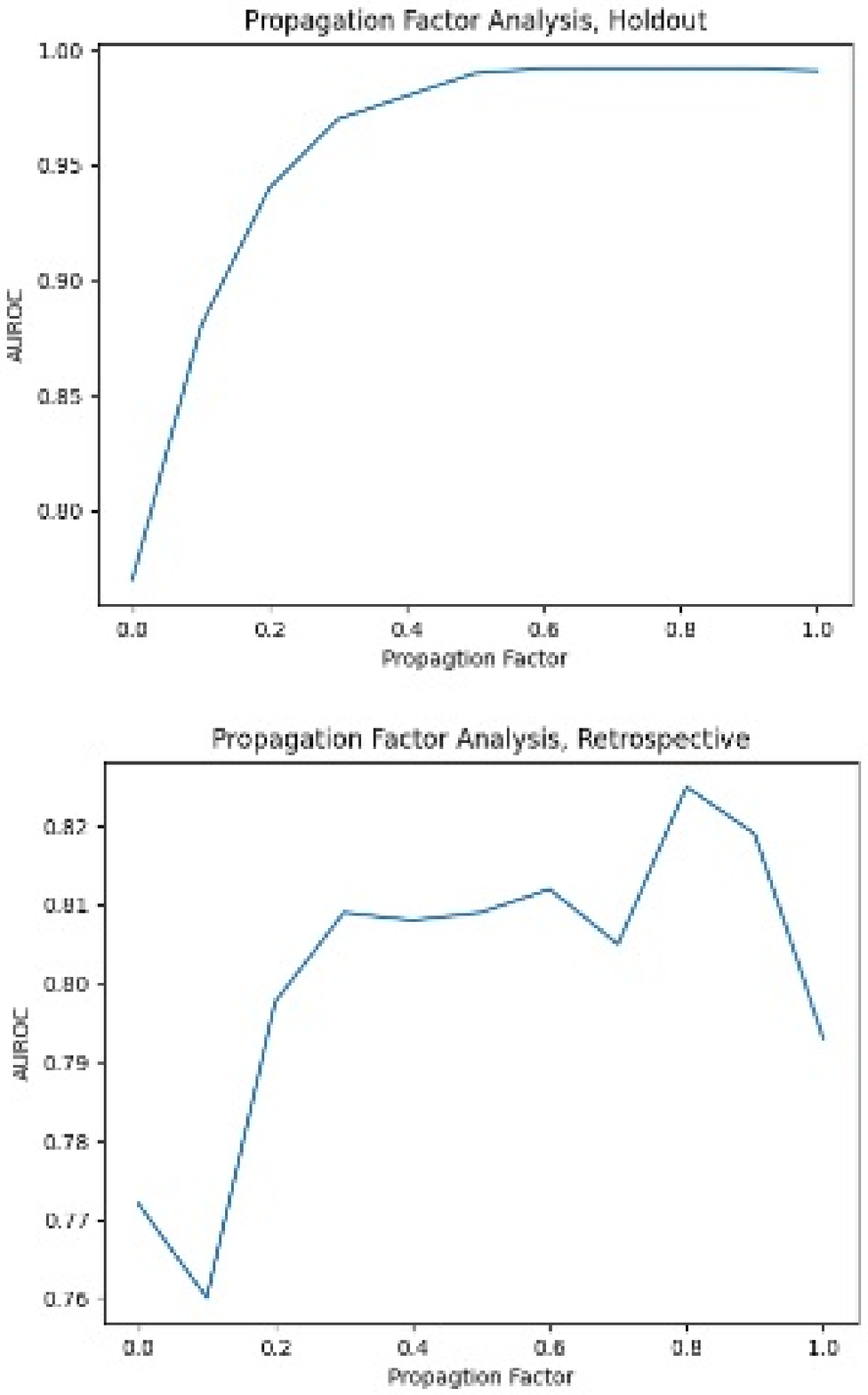}
\caption{{\bf Propagation factor analyses for both evaluations.} \textbf{A)} Holdout evaluation propagation factor evaluation. As can be seen, the value maximizes starting at a propagation factor equal to 0.6 and remains the same until it reaches a propagation factor of $1$ and drops by 0.001 (an insignificant amount). \textbf{B)} Retrospective evaluation propagation factor evaluation. As can be seen, it does not change drastically, and it reaches a maximum value with a propagation factor equal to; this emphasizes the thought that in retrospective evaluation, weights do not change the final results, regardless of whether they are propagated or not, although a little propagation is, as can be seen, appreciated.}
\label{fig5}
\end{figure}

Figure~\ref{fig5} presents the propagation factor analyses for both evaluations. For the holdout evaluation, the score remains almost the same starting from $0.6$; this means that the weights being propagated are very critical for the performance of the holdout evaluation. In contrast, for the retrospective evaluation, the score does not change drastically. However the larger the propagation factor, the larger the AUROC score; this is not the case for AMFP (binary model) which obtained the highest score when the propagation factor is $0.6$. It is then safe to assume that in AMFPMC, the weights that were propagated are of great importance, likely due to the weight balancing required and the fact that there are many interactions with differing numbers of samples, including some with a very small number of samples.

\clearpage
\section*{Contribution}
AMFPMC can be useful in the medical domain to identify unknown interactions. The use of additional information, such as chemical properties like smiles, enzymes, and targets, and chemical substructures, could improve AMFPMC's performance. 
In contrast, most of the recently proposed models have relied solely on chemical properties for drug-drug interaction detection, which means they are not be effective for drugs which the chemical properties are unknown or non-existent (i.e., biologic drugs).

\par
In addition, we have compiled a list of interactions that were identified on DrugBank 5.1.3 and 5.1.6 and made this list publicly available. This list may be helpful for future studies on DDI. This list may also be used by qualified physicians, who are in a position to analyze which interactions are considered harmful and which are considered benign, this knowledge which will enable researchers to focus on those interactions that are considered harmful.
\par
AMFPMC can support large datasets containing many drugs and interactions. In the retrospective evaluation, the scalable model was trained on over $2,000,000$ interactions in just a few minutes and can support even larger datasets. The holdout evaluation showed that on small datasets (with less than $40,000$ interactions) AMFPMC's precision is also obvious, even without the use of any chemical properties. 
\par
By successfully performing multi-class DDI prediction using just graph similarity algorithms, without the need for knowledge on drugs (besides their interactions), we have opened the door to future work on ensemble models combining graph similarity algorithms with other algorithms or properties, and this could enhance existing models. In this research, we did not see the benefit of using such ensemble models, however future research could explore this further, perhaps using different networks, like squeeze-and-excitation networks~\cite{10.1007/s12539-022-00524-0}, which can be used to combine different models and extract one result.

\clearpage
\section*{Appendix}

The tables presented in this section contain the interactions (keyword phrases) identified in the holdout and retrospective analyses. Interactions are listed using their keyword phrases, which represent the main parts of the interaction, without redundant words. For example, the interaction $metabolism \ decreased$ means that $Drug \ A \ decreases \ the \ metabolism \ activities \ of \ Drug \ B $. It is important to note that this does not mean that the interaction between these drugs results in decreased metabolism activities in the patient; rather it means that the metabolism activities of one drug are increased if the drug is taken with another drug. The main difference between the holdout and retrospective analyses in terms of the interactions is that in the holdout evaluation, thane interaction indexed 0 is a real interaction, and in the retrospective evaluation, an interaction indexed 0 indicates that there is no known interaction.
\begin{table}[htp]
\centering
\begin{tabular}{||c c||} 
 \hline
 Index & Keyword phrase  \\ [0.5ex] 
 \hline\hline
 0 & no interaction  \\ 
 1 & increased risk adverse effects  \\
 2 & decreased metabolic activities 
 \\ 
 3 & decreased excretion rate 
  \\
 4 & decreased therapeutic activities 
 \\
 5 &  increased serum concentration  
  \\
 6 &  increased metabolic activities 
  \\
 7 &  increased (CNS depressant) activities 
  \\ 
 8 &  increased QTc prolonging activities 
  \\
 9 &  increased hypotensive activities 
  \\
 10 &  decreased antihypertensive activities 
  \\
 11 &  decreased serum concentration 
  \\
 12 &  increased hypertension 
  \\
 13 & increased therapeutic activities 
  \\
 14 &  increased bleeding 
  \\
 15 &  increased excretion rate 
  \\
 16 &  increased hypoglycemic activities 
  \\
 17 &  increased hyperkalemia  
  \\
 18 &  increased hemorrhage 
  \\
 19 &  increased nephrotoxic activities 
  \\
 20 &  decreased absorption which results in a reduced serum efficacy
  \\
 21 &  increased hypotension 
  \\
 22 &  increased arrhythmogenic activities 
  \\
 23 &  increased thrombogenic activities 
  \\
 24 &  increased bradycardic activities 
  \\
 25 &  increased gastrointestinal activities 
  \\
 26 &  increased tachycardia  \\
 27 &  increased hypertensive  activities 
  \\
 28 &  increased neuroexcitatory activities 
  \\
 29 &  increased renal failure 
  \\
 30 &  increased myopathy 
  \\
 31 &  increased serotonin syndrome 
  \\
 32 &  increased hyperglycemic activities 
  \\
 33 &  increased serotonergic activities 
  \\
 34 &  increased sedative activities 
  \\
 35 &  increased hypokalemic activities 
  \\
 36 & Other (set of uncommon interactions combined)  \\  [1ex] 
 \hline
\end{tabular}

\caption{\bf Indexed interactions for retrospective evaluation. The first column contains the index assigned to the interaction inside the code, and the second column contains the keyword phrase extracted for simpler calculation (redundant words and words that are redundant because the problem is symmetric were removed). The keyword sentences were extracted from full interaction sentences; "other" interactions can be seen in our comprehensive extraction of interactions~\cite{interactionsExtractions}.}
\label{table:1}
\end{table}
\clearpage

\begin{table}[htp]
\centering
\begin{tabular}{||c c||} 
 \hline
 Index & Keyword Sentence  \\ [0.5ex] 
 \hline\hline
 0 & decreased metabolism   \\ 
 1 & increased risk adverse effects  \\
 2 & increased serum  concentration
 \\ 
 3 & decreased serum concentration
  \\
 4 & decreased therapeutic efficacy 
 \\
 5 &  increased (CNS depressant) 
  \\
 6 & increased QTc-prolonging activities  \\
 7 &  increased hypotensive activities 
  \\ 
 8 &  increased metabolism 
  \\
 9 &  decreased hypoglycemic activities 
  \\
 10 & decreased anti hypertensive activities 
  \\
 11 &  increased anticoagulant activities 
  \\
 12 & increased bradycardic activities 
  \\
 13 &  increased serotonergic activities 
  \\
 14 &  increased therapeutic efficacy 
  \\
 15 &  increased hypokalemic activities 
  \\
 16 & increased  orthostatic hypotensive activities 
  \\
 17 & decreased cardiotoxic activities 
  \\
 18 &  decreased excretion rate 
  \\
 19 &  increased atrioventricular blocking activities 
  \\
 20 &  decreased sedative activties 
  \\
 21 & increased tachycardic activities 
  \\
 22 &  increased hypertensive and vasoconstricting activities 
  \\
 23 &  increased QTc prolongation 
  \\
 24 &  increased anti hypertensive activities 
  \\
 25 &  increased arrhythmogenic activities  \\
 26 &  increased cardiotoxic activities 
  \\
 27 &  increased hypotension 
  \\
 28 &  decreased bronchodilatory activities 
  \\
 29 &  increased hyperkalemic activities 
  \\
 30 & increased nephrotoxic activities 
  \\
 31 &  increased hypertensive activities  \\
 32 &  increased vasoconstricting activities 
  \\
 33 &  increased neuroexcitatory activities 
  \\
 34 &  increased fluid retaining activities 
  \\
  35 &  decreased serum active metabolites 
  \\
  36 &  increased immunosuppressive activities 
  \\
  37 &  increased bleeding 
  \\
  38 &  decreasedstimulatory activities 
  \\
  39 & increased myelosuppression 
  \\
  40 &  decreased vasoconstricting activities 
  \\
  41 &  increased thrombogenic activities 
  \\
  42 &  increased analgesic activities 
  \\
  43 & increased anticholinergic activities 
  \\
  44 &  increased absorption 
  \\
  45 &  increased sedation and somnolence  
  \\
  46 &  increased rhabdomyolysis 
  \\
  47 &  increased hyperkalemia 
  \\
  48 &  increased hepatotoxic 
  \\
  49 & increased  myopathic rhabdomyolysis activities 
  \\
  50 &  decreased absorption
  \\
  51 & increased  hyponatremic activities 
  \\
  52 &  increased vasopressor activities 
  \\
  53 &  increased excretion rate 
  \\
  54 &  decreased neuromuscular blocking activities 
  \\
  55 & increased hypersensitivity  \\
  56 & increased hyperglycemic activities  \\
  57 & increased hypocalcemic activities  \\
  58 & increased myelosuppressive activities  \\
  59 & increased vasodilatory activities  \\ [1ex]
 \hline
\end{tabular}

\caption{\bf Indexed interactions for holdout evaluation. The first column contains the index assigned to the interaction inside the code, and the second column contains the keyword phrase extracted for simpler calculation (redundant words and words that are redundant because the problem is symmetric were removed). The keyword sentences were extracted from full interaction sentences; "other" interactions can be seen in our comprehensive extraction of interactions~\cite{interactionsExtractions}.}

\label{table:1}
\end{table}
\clearpage

\bibliography{DDIP}{}

\end{document}